\documentclass[conference]{IEEEtran}
\IEEEoverridecommandlockouts
\usepackage{amsmath,amsfonts,amssymb} 
\usepackage{array}
 
\usepackage[caption=false,font=footnotesize,labelfont=sf,textfont=sf]{subfig}
\usepackage{textcomp} 
\usepackage{url} 
\usepackage{graphicx} 
\usepackage[colorlinks=true, linkcolor=blue, citecolor=blue, urlcolor=blue, bookmarks=false, pdfpagemode=UseNone]{hyperref}
\usepackage{cite} 
\usepackage{makecell} 
\usepackage{enumitem} 
\usepackage[linesnumbered,ruled,boxed]{algorithm2e}

\usepackage{color} 
\usepackage{booktabs}
\usepackage{stfloats}
\usepackage{multirow}
\usepackage[letterpaper, top=1in, left=0.75in, right=0.75in, bottom=0.75in]{geometry}

\begin{document}
	
	\title{Learning a System-Level Surrogate for Hydraulic Excavators: A Simulation-to-Real LSTM Approach}

	\author{
		Shuai Wang$^{1}$,
		Shen Wang$^{1*}$,
		Qiang Wang$^{1}$,
		Muguo Du$^{2}$,
		Donghai Shi$^{2}$,
		Chenyu Wang$^{1}$,
		Xiaofeng Tao$^{1}$
		\thanks{This work is supported by the National Natural Science Foundations of China under Grant 62203062, and partly by the Fundamental Research Funds for the Central Universities under Grants 2242022k60006 and 2024RC08, and BUPT-Yaowutech cooperation. $^{*}$Corresponding author: Shen Wang (shen.wang@bupt.edu.cn). 
			
		$^{1}$Shuai Wang, Shen Wang, Qiang Wang, Chenyu Wang, and Xiaofeng Tao are with the National Engineering Research Center of Mobile Network Technologies, Beijing University of Posts and Telecommunications, Beijing 100876, China (e-mails: {shuai.wang, shen.wang, wangq, wangchenyu, taoxf}@bupt.edu.cn).}%
		\thanks{$^{2}$Muguo Du and Donghai Shi are with the Yaowu (Shenzhen) Technology Co., Ltd., Shenzhen 518052, China (e-mails: {dumuguo, shidonghai}@yaowutech.com).}%
		
	}
	
	\maketitle
	
	\begin{abstract}
		Developing autonomous hydraulic excavators is constrained by limited access to physical machines and the high cost of real-world experimentation. This paper proposes a simulation-to-real framework for learning a system-level digital surrogate using Long Short-Term Memory (LSTM) networks. Instead of modeling internal dynamics, the excavator is treated as an input–output operator, and the surrogate is trained to reproduce its closed-loop behavior under identical control inputs. The approach is first validated in a MuJoCo simulation environment and then transferred to a real excavator. To address measurement inconsistencies in real-world data, a consistency-aware state estimation method based on adaptive Kalman filtering is introduced. Experimental results demonstrate that the learned surrogate achieves high fidelity in both angular velocity and long-horizon trajectory reproduction under closed-loop autoregressive evaluation. These results confirm that the proposed model can serve as a drop-in surrogate for both simulation and physical systems, enabling scalable and efficient development of excavation automation algorithms.
	\end{abstract}


	\section{Introduction}\label{sec:introduction}
	
	Hydraulic excavators play a critical role in construction, mining, and disaster response, where their operation directly impacts efficiency and safety~\cite{zhang_autonomous_2021}. With the rapid advancement of robotics and reinforcement learning, there is growing interest in enabling autonomous or semi-autonomous excavation systems~\cite{egli_general_2022}. However, the development of such systems remains fundamentally constrained by a practical bottleneck: the limited availability of physical machines, the high cost of experimentation, and the low efficiency of iterative development~\cite{hoffmannDatadrivenModelingControl}.
	
	Unlike typical robotic platforms, hydraulic excavators are large-scale, safety-critical systems with complex actuation mechanisms. In most research settings, only a single machine is available, making it difficult to support iterative algorithm development, large-scale data collection, or parallel experimentation. Moreover, direct deployment of unvalidated algorithms on physical machines is both time-consuming and risky. These limitations highlight the urgent need for a reliable digital surrogate that can replace the physical system during development and testing.

	Constructing such a surrogate, however, is highly challenging. The hydraulic excavator exhibits strong nonlinear dynamics induced by hydraulic actuation, along with inherent time-delay effects arising from valve response and fluid transmission~\cite{leePrecisionMotionControl2022}. These characteristics are difficult to capture using explicit physics-based models due to complex parameter dependencies and unmodeled internal dynamics.  Although physics-based modeling and system identification methods have been extensively studied~\cite{narendraIdentificationControlDynamical1990,leeRobotModelIdentification2024}, they often require precise parameterization and fail to capture unmodeled dynamics in real-world excavators.
	
	From a learning perspective, the challenge is further compounded by the ambiguity of the modeling target. Traditional system identification methods focus on learning explicit dynamics models of the form $\ \boldsymbol{x}_{t+1}=f(\boldsymbol{x}_t,\boldsymbol{u}_t)$. In contrast, practical excavator systems behave as complex input-output operators, where the goal is to learn the end-to-end mapping from control commands to observed system responses. This shift fundamentally changes the problem from component-level dynamics modeling to system-level behavior approximation. Recent data-driven approaches, particularly those based on recurrent neural networks such as LSTM, have shown strong capability in modeling nonlinear dynamical systems~\cite{heindelDatadrivenApproachApproximating2022,wenLSTMbasedAdaptiveRobust2023,maDataDrivenMultistepNonlinear2024,hoffmannModelingWeaklyinstrumentedExcavator}. However, existing works primarily focus on modeling isolated subsystems or rely on relatively clean and well-instrumented data, leaving the system-level surrogate problem underexplored.
	
	Another major challenge lies in the discrepancy between simulation and reality. While physics engines such as MuJoCo can accurately model rigid-body dynamics, they typically do not capture hydraulic effects, delays, or real-world disturbances. Consequently, models trained purely in simulation often suffer from a significant sim-to-real gap~\cite{aoshimaExaminingSimulationtorealityGap2025}.
	
	Finally, data quality issues in real systems further complicate the learning process. Sensors on hydraulic excavators are highly susceptible to vibration and disturbance~\cite{hoffmannModelingWeaklyinstrumentedExcavator}, leading to inconsistencies between angle measurements and angular velocity feedback. Such inconsistencies violate basic kinematic constraints and can severely degrade the performance of data-driven models if not properly addressed~\cite{livierisSmoothingStationarityEnforcement2021}.
	
	To address these challenges, this paper proposes a simulation-to-real learning paradigm for constructing system-level digital surrogates of hydraulic excavators using Long Short-Term Memory (LSTM) networks. The key idea is to treat the entire excavator pipeline as a black-box operator and learn its input-output behavior directly from data, rather than explicitly modeling individual subsystems. The framework is implemented in a staged manner.

	First, a MuJoCo-based simulation environment is employed as an idealized setting to validate whether an LSTM model can learn a consistent input-output mapping under well-controlled conditions. The simulator provides clean, low-noise observations and captures rigid-body dynamics, but does not model hydraulic effects or real-world disturbances. Therefore, validation in MuJoCo primarily establishes the methodological correctness of the proposed surrogate modeling approach.
	
	The same methodology is then transferred to a digitally retrofitted real excavator, where the system exhibits significantly higher complexity due to unmodeled hydraulic dynamics, time delays, and sensor imperfections caused by vibration and drift. To address these challenges, a consistency-aware state estimation method based on adaptive Kalman filtering~\cite{rodriguezEnhancedAdaptiveKalman2025} is introduced to ensure reliable training data. Validation on the physical system further demonstrates the realism of the learned surrogate, confirming its ability to capture the full system behavior beyond the simplified simulation model.

The main contributions of this paper are summarized as follows: 
\begin{itemize}
	\item We propose a data-driven paradigm that models the hydraulic excavator as an input–output operator, enabling end-to-end learning of system behavior without requiring explicit dynamics modeling.
	
	\item We introduce a closed-loop behavioral equivalence evaluation protocol and demonstrate, through both simulation and real-world experiments, that the learned LSTM surrogate can faithfully reproduce system behavior under identical control inputs.
	
	\item We develop an adaptive Kalman filtering method to enforce kinematic consistency in noisy sensor data, which significantly improves the stability and reliability of surrogate learning on physical systems.
\end{itemize}

The remainder of this paper is organized as follows. 
Section~\ref{sec:system description} introduces the excavator platform and formulates the surrogate modeling problem as an input–output operator. 
Section~\ref{sec:framework} presents the staged simulation-to-real framework and the LSTM-based surrogate model. 
Section~\ref{sec:kalman} describes the consistency-aware state estimation method for real-world data. 
Section~\ref{sec:experiments} provides closed-loop experimental validation and analysis. 
Section~\ref{sec:conclusions} concludes the paper.

		\section{System Description}\label{sec:system description}
To enable control substitutability, a unified hardware--software interface is first established, allowing the excavator to be treated as a programmable and observable system. 
		
		\subsection{Physical Excavator Platform and Digital Retrofitting}\label{sec:PhysicalExcavator}

The platform used in this study is a digitally retrofitted SY750H commercial hydraulic excavator equipped with an electronically controlled hydraulic actuation system; see Figure~\ref{fig:framework}. This system enables direct reception of digital control commands at the joint level. Four primary joints---swing, boom, arm, and bucket---are independently controlled through a unified interface with normalized inputs:
\begin{equation}
	\boldsymbol{u}_t = [u_{0,t}, u_{1,t}, u_{2,t}, u_{3,t}]^\top \in [-1000,1000]^4.
\end{equation}

By abstracting low-level valve actuation into a standardized API, the system exposes a hardware-independent control interface, effectively decoupling algorithm development from hydraulic implementation details.

For state acquisition, each joint is instrumented with a dynamic inclinometer (HDA437TC3, RION Technology), providing measurements of joint angles and angular velocities; see Figure~\ref{fig:framework}. These sensors are mounted directly on the mechanical structure to capture real-time kinematic responses under both no-load and operational conditions.
		
\subsection{Problem Formulation}
		
The goal of this work is to construct a surrogate model that reproduces the dynamic behavior of the excavator under given control inputs. Unlike conventional system identification~\cite{PetrovićMathematicalmodelling2022}, which assumes an explicit state-transition model of the form $\boldsymbol{x}_{t+1}=f(\boldsymbol{x}_t,\boldsymbol{u}_t)$, the excavator system considered here is dominated by nonlinear hydraulic actuation and intrinsic time-delay effects. These characteristics make it difficult to explicitly model internal state transitions, and instead motivate learning a system-level input-output operator.

Accordingly, the system is formulated as an operator, i.e,
$
	\boldsymbol{x}_t = \mathcal{F}(\boldsymbol{u}_{1:t}),
$
where $\mathcal{F}$ implicitly captures the combined effects of nonlinear dynamics and delays. In practice, due to partial observability, the operator is approximated using a finite history window, that is,
\begin{equation}\label{eq:surrogate}
	\tilde{\boldsymbol{x}}_{t+1}=\Phi_\theta(\boldsymbol{x}_{t-W+1:t},\boldsymbol{u}_{t-W+1:t}),
\end{equation}
where $\Phi_\theta$ is a parameterized model and $W$ denotes the temporal window length. The objective is not only accurate one-step prediction, but more importantly, long-horizon consistency under autoregressive rollout, such that the learned surrogate can serve as a functional replacement of the physical system in closed-loop settings.
	
	\subsection{From Physical System to Surrogate Modeling}
	
The above formulation highlights that the excavator should be treated as a system-level input-output operator rather than a collection of explicitly modeled subsystems. Since the primary objective in control and algorithm development is to reproduce the overall response behavior, we adopt a data-driven surrogate modeling approach that directly learns this operator from observations.

Based on this perspective, we design a staged framework to construct and validate the surrogate model, as described in the next section.
	
\section{Staged Digital Surrogate Framework and LSTM-based Modeling}\label{sec:framework}

\subsection{Overall Pipeline}
	
\begin{figure*}[!htbp] 
		\centering
		\includegraphics[width=0.98\textwidth]{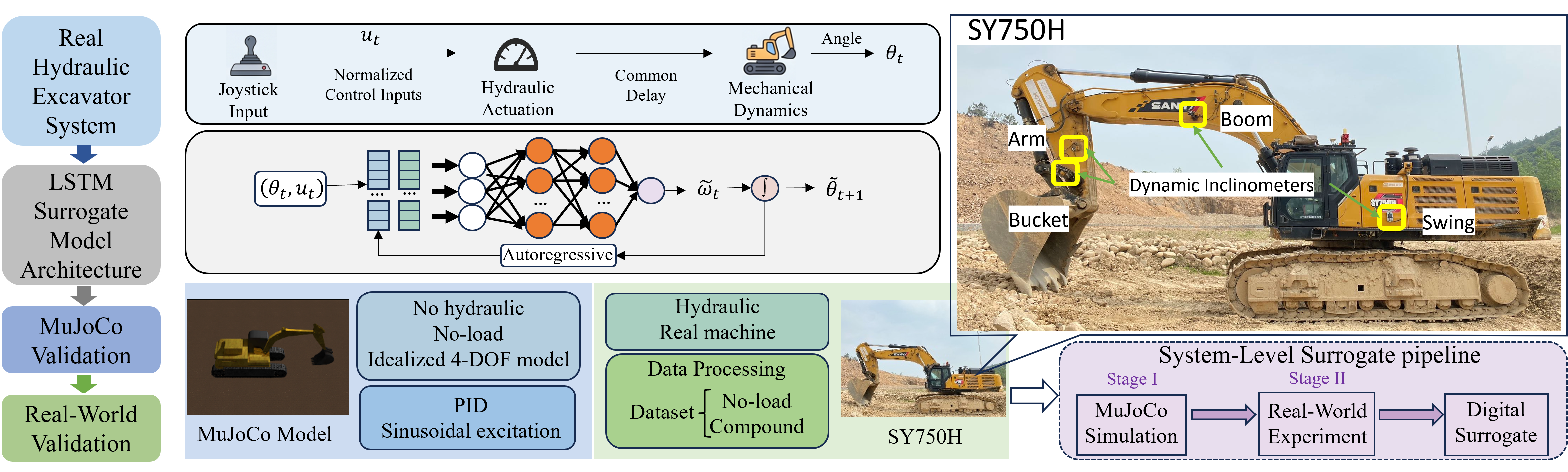}
		\caption{Staged digital surrogate framework illustrating operator learning and simulation-to-real transfer.}
		\label{fig:framework}
		\vspace{-1em}
	\end{figure*}

	As illustrated in Fig.~\ref{fig:framework}, the proposed framework consists of two stages. First, a MuJoCo-based simulation environment is constructed to validate whether the operator learning approach can reproduce system behavior under ideal conditions. Then, the same methodology is transferred to the physical excavator, where real-world data is used to learn a surrogate that captures practical dynamics.
	
	\subsubsection{Stage I: MuJoCo Validation}
	
	A four-degree-of-freedom excavator model is constructed in MuJoCo based on geometric and inertial properties. The simulation provides a noise-free and fully observable environment to verify whether the learned surrogate can reproduce system behavior under controlled conditions. To approximate actuation delay, a fixed latency of 50\,ms is introduced, consistent with reported delay ranges in hydraulic excavators~\cite{yamamotoPositionControllerHydraulic2025}. This stage focuses on validating the correctness of the operator learning paradigm.
	
	\subsubsection{Stage II: Deployment on Real-World Physical System}
	
	Compared to simulation, the real excavator introduces additional complexities, including hydraulic nonlinearities, dead zones, hysteresis, and sensor noise. Therefore, before training the surrogate model, data consistency processing is applied to ensure reliable state estimation. This stage evaluates whether the learned operator can faithfully reproduce real-world system behavior, thereby validating its practical applicability.
	 
	 	\subsection{LSTM-Based Surrogate Modeling}\label{sec:lstm}

	 The surrogate model aims to approximate the system-level operator defined in Eq.~\eqref{eq:surrogate}. In this work, a Long Short-Term Memory (LSTM) network is adopted as a sequence modeling tool to capture temporal dependencies in the input-output behavior of the system.

\subsubsection{Input and Output Definition}

The model input is a 9-dimensional feature vector composed of the joint states \[
\boldsymbol{x}_t = [\sin\theta_{0,t}, \cos\theta_{0,t}, \theta_{1,t}, \theta_{2,t}, \theta_{3,t}]^\top,
\] and normalized control command vector $\boldsymbol{u}_t$ at the current time step, where $\theta_{0,t}$, $\theta_{1,t}$, $\theta_{2,t}$, and $\theta_{3,t}$ correspond to the angles of the swing, boom, arm, and bucket, respectively. Note that, $\theta_0$ is encoded using $\sin/\cos$ to ensure continuity across periodic boundaries. The output is the joint angular velocity
$
	\tilde{\boldsymbol{y}}_t = [\tilde{\omega}_{0,t}, \tilde{\omega}_{1,t}, \tilde{\omega}_{2,t}, \tilde{\omega}_{3,t}]^\top.
$

\subsubsection{Network Structure and Training}
A two-layer stacked LSTM with hidden dimension 128 is used.  The gating mechanism enables the network to adaptively capture dynamic dependencies across various time scales~\cite{rostamijavananiDatadrivenIdentificationNonlinear2025}. The LSTM output is mapped to angular velocity predictions through a fully connected layer with an intermediate ReLU activation and a linear output layer. The input sequence is constructed using a sliding window of length $W=30$; see Eq.~\eqref{eq:surrogate}. Model parameters are optimized using Adam with learning rate $10^{-3}$ and adaptive scheduling.

To ensure long-horizon consistency, a multi-step rollout training strategy is adopted. The loss function is defined as
$
	\mathcal{L} = \mathcal{L}_{\text{vel}} + \lambda_{\text{bias}} \cdot \mathcal{L}_{\text{bias}} ,
$
where $\mathcal{L}_{\text{vel}}$ is the Smooth L1 loss over the $H$-step predicted velocity sequence, measures prediction error and
$
	\mathcal{L}_{\text{vel}} = \frac{1}{H}\sum_{h=1}^{H}\text{SmoothL1}(\tilde{\boldsymbol{y}}_{t+h},\;\boldsymbol{y}_{t+h}),
$
with $\text{SmoothL1}(a,b) = \frac{1}{2}(a-b)^2$ if $|a-b|<1$, and $|a-b|-\frac{1}{2}$ otherwise. $\mathcal{L}_{\text{bias}}$ penalizes the mean prediction bias over the rollout horizon, and
$
	\mathcal{L}_{\text{bias}} = \left\| \frac{1}{H}\sum_{h=1}^{H}\tilde{\boldsymbol{y}}_{t+h} - \frac{1}{H}\sum_{h=1}^{H}\boldsymbol{y}_{t+h} \right\|^2.
$ This design suppresses systematic drift during autoregressive rollout and ensures that the learned surrogate maintains behavioral consistency over long horizons. When $H=1$, the formulation reduces to standard one-step training.

	\section{Consistency-Aware State Estimation}\label{sec:kalman}
	
	In the physical excavator, the acquired joint states exhibit significant inconsistency due to sensor imperfections. Angle measurements (from inclinometers; see Section \ref{sec:PhysicalExcavator} for details) may contain transient jumps caused by vibration or electromagnetic interference, while angular velocity signals (from inertial sensors) suffer from cumulative drift over time. As a result, the raw signals $(\theta, \omega)$ often violate kinematic consistency, which degrades the quality of the learned surrogate model. To address this issue, we employ an adaptive Kalman filtering scheme with consistency detection to fuse angle and angular velocity measurements. The objective is not to improve estimation accuracy per se, but to enforce physically consistent training data for system-level surrogate modeling~\cite{livierisSmoothingStationarityEnforcement2021}.

	\subsection{Consistency Detection and Adaptive Fusion}
	
	For each joint, the state is defined as $\boldsymbol{s}_t = [\theta_t, \omega_t]^\top$, with a constant-velocity prediction model $
	\boldsymbol{s}_t = \boldsymbol{A}\boldsymbol{s}_{t-1} + \boldsymbol{w}_t, \quad \boldsymbol{A} = \begin{bmatrix} 1 & T_s \\ 0 & 1 \end{bmatrix},
	$ where $T_s$ is the sampling period, and $\boldsymbol{w}_t$ represents the process noise with a covariance matrix denoted by $\boldsymbol{Q} = \text{diag}(\boldsymbol{Q}_\theta, \boldsymbol{Q}_{\omega})$, in which $\boldsymbol{Q}_\theta$ and $\boldsymbol{Q}_{\omega}$ are the process noise variances for the angle and angular velocity channels, respectively. The observation model is given by
	$
	z_t = \boldsymbol{H}\boldsymbol{s}_t + v_t, \quad \boldsymbol{H} = \begin{bmatrix} 1 & 0 \end{bmatrix},
	$
	where $z_t$ denotes the measured joint angle and $v_t$ is the observation noise with a covariance of $R_t$. 
	
	To handle unreliable measurements, a binary consistency flag $c_t \in \{0,1\}$ is introduced based on two criteria: (i) {Jump detection:}  $\lvert \theta_{t} - \theta_{t-1} \rvert > \tau_{\text{jump}}$; (ii) {Consistency check:} deviation between $\theta_{t}$ and the angle $\theta^{intg}_{t}$ reconstructed from integrating $\omega$ over a short window. The observation noise is then adaptively adjusted as
	\begin{equation}
		R_t =
		\begin{cases}
			R_{\text{trust}}, & c_t = 1, \\
			R_{\text{untrust}}, & c_t = 0,
		\end{cases}
	\end{equation}
	where $R_{\text{untrust}} \gg R_{\text{trust}}$. When the measurement is unreliable, the filter relies more on angular velocity integration; otherwise, it anchors to the angle observation. The resulting consistency-aware  adaptive Kalman filtering procedure is summarized in Algorithm~\ref{alg:AKF}.

	\subsection{Impact on Surrogate Modeling}
	
	After fusion, the angle signal is smooth and free of discontinuities, and the angular velocity is recovered via numerical differentiation:
	\begin{equation}
		\hat{\omega}_{t} = \frac{\hat{\theta}_{t} - \hat{\theta}_{t-1}}{T_s}.
	\end{equation}
	The fused states $(\hat{\theta}_t, \hat{\omega}_t)$ replace the raw sensor readings in the LSTM input vector $\boldsymbol{x}_t$ defined in Eq.~\eqref{eq:surrogate}, ensuring that the surrogate model is trained on kinematically consistent data. This prevents the model from fitting contradictory signals and significantly improves long-horizon autoregressive stability in real-world deployment.

	\begin{algorithm}[h]
		\caption{Adaptive State Estimation Algorithm Based on Consistency Detection}
		\label{alg:AKF}
		\KwIn{Measurement sequence $\{\theta_{t}, \omega_t\}$, thresholds $\tau_{\text{jump}}, \tau_{\text{cons}}$, weights $R_{\text{trust}}, R_{\text{untrust}}$}
		\KwOut{Fused state $\hat{\boldsymbol{s}}_t = [\hat{\theta}_t, \hat{\omega}_t]^\top$, trusted flag $c_t$}
		Initialize $\hat{\boldsymbol{s}}_0, \boldsymbol{P}_0$\;
		\For{t = 1 \text{ to } N}{
			$\hat{\boldsymbol{s}}_{t|t-1} = \boldsymbol{A}\hat{\boldsymbol{s}}_{t-1}$\;
			$\boldsymbol{P}_{t|t-1} = \boldsymbol{A}\boldsymbol{P}_{t-1}\boldsymbol{A}^\top + \boldsymbol{Q}$\;

			Calculate instantaneous jump $\delta_t = \lvert \theta_{t} - \theta_{t-1} \rvert$\;
			Calculate window consistency error $\epsilon_t = \text{RMS}(\theta_{t} - \theta^{intg}_{t}, w)$\;

			\eIf{$\delta_t < \tau_{\text{jump}} \land \epsilon_t < \tau_{\text{cons}}$}{
				$c_t = 1, R_t = R_{\text{trust}}$\;
			}{
				$c_t = 0, R_t = R_{\text{untrust}}$\;
			}

			$\boldsymbol{K}_t = \boldsymbol{P}_{t|t-1}\boldsymbol{H}^\top(\boldsymbol{H}\boldsymbol{P}_{t|t-1}\boldsymbol{H}^\top + R_t)^{-1}$\;
			$\hat{\boldsymbol{s}}_t = \hat{\boldsymbol{s}}_{t|t-1} + \boldsymbol{K}_t(z_t - \boldsymbol{H}\hat{\boldsymbol{s}}_{t|t-1})$\;
			$\boldsymbol{P}_t = (\boldsymbol{I} - \boldsymbol{K}_t\boldsymbol{H})\boldsymbol{P}_{t|t-1}$\;
		}
	 
	\end{algorithm}

	\section{Experimental Validation and Analysis}\label{sec:experiments}

	\subsection{Evaluation Protocol}
	
	The surrogate is evaluated via \emph{closed-loop behavioral equivalence} (CBE), where the model evolves autoregressively under identical control inputs without external correction. Performance is measured using Root Mean Square Error (RMSE), Mean Absolute Error (MAE), and the coefficient of determination ($R^2$) on both angular velocity and integrated angle trajectories, capturing short-term accuracy and long-horizon consistency.
	
	As summarized in Table~\ref{tab:exp_overview}, Stage I (MuJoCo) validates the modeling paradigm under ideal conditions, while Stage II evaluates transfer to the physical system (SY750H excavator in Figure.~\ref{fig:framework}) under two scenarios: (i) \emph{no-load condition}, corresponding to free-space joint motion without external load, reflecting intrinsic system dynamics, and (ii) \emph{compound operation condition}, involving a full working cycle (digging, lifting, swinging, and dumping), where the system is subject to load variations and stronger nonlinear effects.
	\begin{table}[htbp]
		\centering
		\caption{Overview of experimental configurations.}
		\label{tab:exp_overview}
		\begin{tabular}{l c c c}
			\toprule
			\textit{Parameter} & \textit{\makecell{Stage I:\\MuJoCo\\Simulation}} & \textit{\makecell{Stage II:\\Real-World \\No-load}} & \textit{\makecell{Stage II:\\Real-World \\Compound}} \\
			\midrule
			$f_s$ (Hz)$^\ast$  & 20 & 20 & 10 \\
			Training sets & 12 & 34 & 33 \\
			Samples (k)$^\dagger$ & 63.6k & 650k & 84k \\
			$H$ (rollout) & 1 & 1 & 10 \\
			$\lambda_{\text{bias}}$ & 0 & 0 & 0.3 \\
			Test time (s) & 90 & 382.85 & 295 \\
			Test steps & 1800 & 7657 & 2950 \\
			\bottomrule
			\multicolumn{4}{l}{\footnotesize $^\ast$\,$f_s = 1/T_s$, where $T_s$ is the sampling period. } \\
			\multicolumn{4}{l}{\footnotesize $^\dagger$\,k denotes $\times 10^3$ samples.}
		\end{tabular}
		\vspace{-1em}
	\end{table}

	\begin{figure}[!htbp] 
		\centering
		\includegraphics[width=0.48\textwidth]{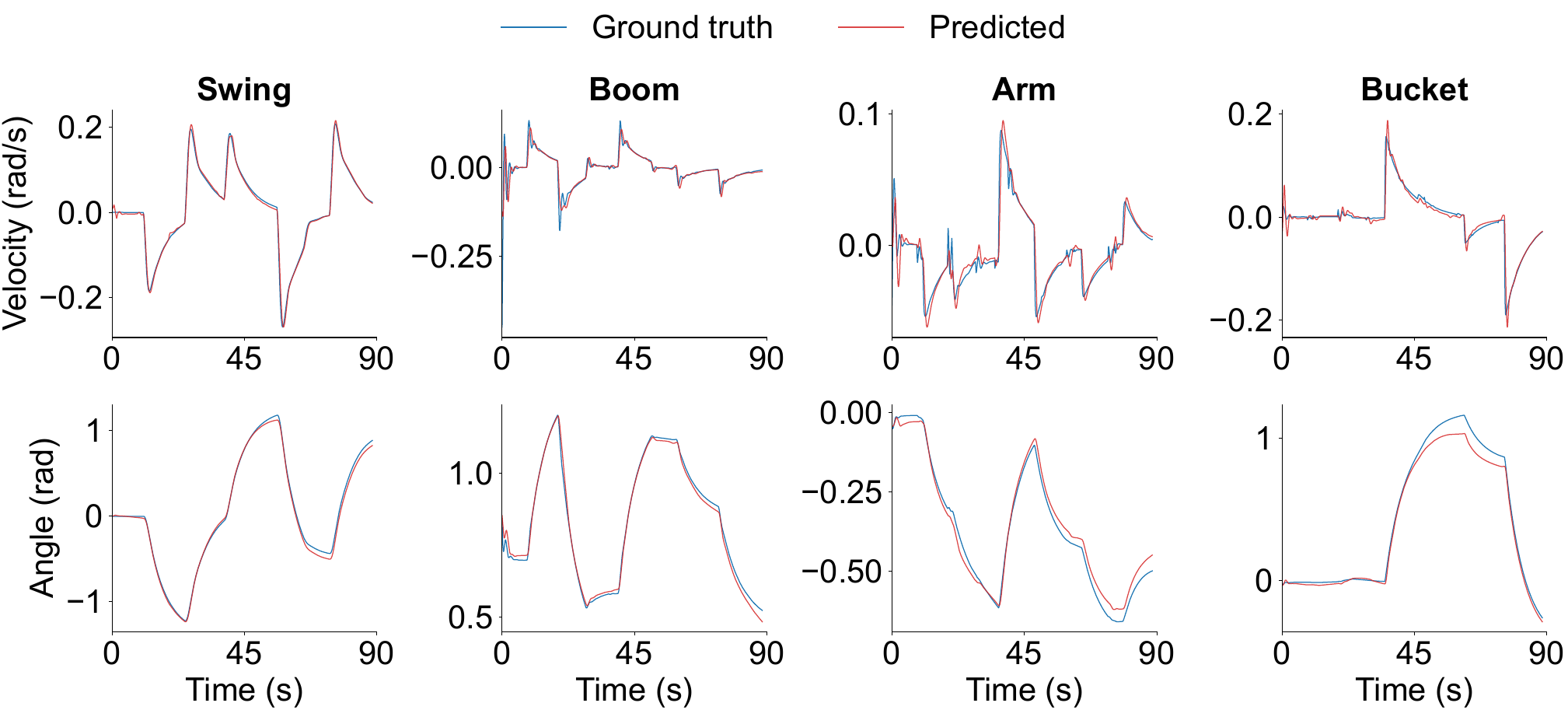}
		\caption{Closed-loop validation in simulation: comparison between the LSTM free-run trajectory and the MuJoCo ground-truth trajectory. The upper row shows angular velocity and the lower row shows integrated angle trajectories.}
		\label{fig:autoreg_compare_mujoco}
		\vspace{-1em}
	\end{figure}  
	
		\begin{table}[htbp]
		\centering
		\caption{Stage I: Closed-loop autoregressive validation metrics in MuJoCo simulation.}
		\label{tab:autoreg_mujoco_metrics}
		\begin{tabular}{c c c c c}
			\toprule
			\textit{Type} & \textit{Joint} & \textit{RMSE} & \textit{MAE} & \textit{$R^2$} \\
			\midrule
			\multirow{4}{*}{\textit{\makecell{Angular \\velocity\\ $\omega$}}}
			& Swing & 0.0074 & 0.0052 & 0.9940 \\
			& Boom        & 0.0219 & 0.0083 & 0.7726 \\
			& Arm         & 0.0087 & 0.0051 & 0.8965 \\
			& Bucket      & 0.0122 & 0.0059 & 0.9424 \\
			\midrule
			\multirow{4}{*}{\textit{\makecell{Angle\\ trajectory\\ $\theta$ }}}
			& Swing & 0.0380 & 0.0303 & 0.9967 \\
			& Boom        & 0.0179 & 0.0138 & 0.9927 \\
			& Arm         & 0.0280 & 0.0249 & 0.9809 \\
			& Bucket      & 0.0585 & 0.0436 & 0.9854 \\
			\bottomrule
		\end{tabular}
	\end{table}
	
	\subsection{Stage I: MuJoCo Simulation Validation}
	
	A four-DoF excavator model is constructed in MuJoCo (v3.3.7)~\cite{todorovMuJoCo2012} using geometric and inertial parameters. A fixed delay of 50\,ms is introduced to approximate actuation latency. The training dataset consists of 12 sinusoidal excitation trajectories (63.6k samples at 20\,Hz). Evaluation is performed on unseen step-input trajectories to assess generalization.
	
	Fig.~\ref{fig:autoreg_compare_mujoco} shows the closed-loop comparison between the LSTM surrogate and the MuJoCo simulator. Despite fully autoregressive rollout over 1800 steps (90\,s), the predicted trajectories remain closely aligned with the ground truth. Quantitative results in Table~\ref{tab:autoreg_mujoco_metrics} indicate high fidelity across all joints, with $R^2$ exceeding 0.94 for angle trajectories. These results demonstrate that the LSTM successfully captures the system-level input-output behavior in an idealized environment and can serve as a functionally equivalent surrogate of the simulator.
	
	\subsection{Stage II: Real-World Validation}
	
	\subsubsection{Data Preparation}
	
\begin{figure}[htbp]
		\centering
		\includegraphics[width=0.7\columnwidth]{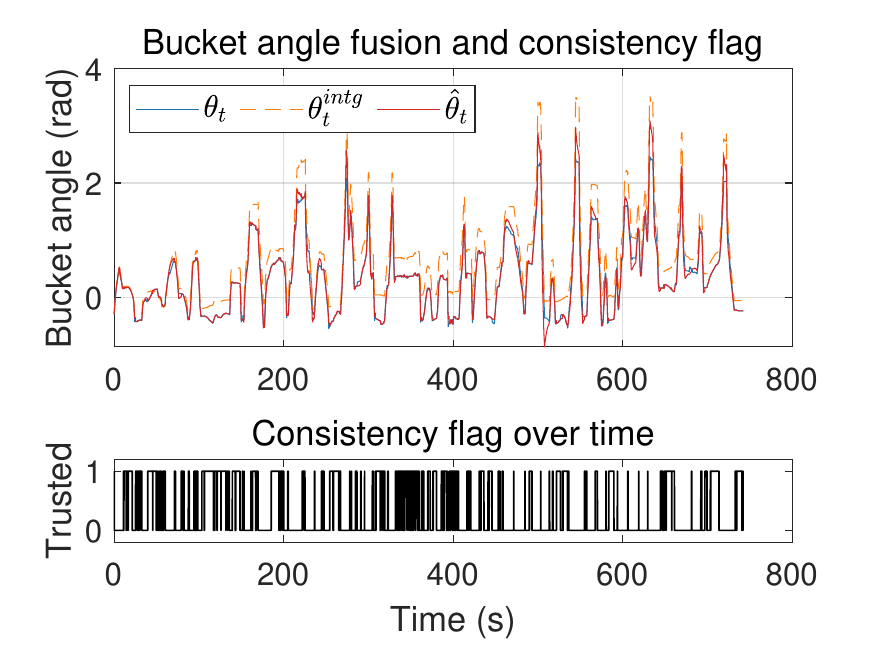}
		\caption{Kalman filtering fusion for the bucket joint. The blue line represents the raw angle measurement $\theta_t$, the orange line represents the reconstructed angle $\theta^{intg}_t$ obtained by integrating the angular velocity $\omega_t$, and the red line represents the fused estimate $\hat{\theta}_t$ produced by Algorithm~\ref{alg:AKF}.}
		\label{fig:bucketanglefusion}
	\end{figure}

	In the physical system, raw angle and angular velocity measurements exhibit inconsistencies due to sensor noise, vibration, and drift. To address this issue, the adaptive Kalman filtering method (Algorithm~\ref{alg:AKF}) described in Section~\ref{sec:kalman} is applied to enforce kinematic consistency. Fig.~\ref{fig:bucketanglefusion} illustrates a representative fusion result (the bucket joint). The RMSE between the fused and raw measured angles for the swing, boom, arm, and bucket joints are 0.0037\,rad, 0.0041\,rad, 0.0119\,rad, and 0.1093\,rad, respectively. The processed data are then used for training and evaluation.
	
	\subsubsection{No-Load Validation}
	
	A total of 34 datasets (approximately 650k samples) are collected under no-load conditions. The surrogate model is evaluated over 7657 autoregressive steps (382.85\,s at 20\,Hz) without external correction (Table~\ref{tab:exp_overview}).
	
	Fig.~\ref{fig:autoreg_real} presents the comparison between the LSTM surrogate and the physical excavator. The model reproduces both angular velocity responses and integrated angle trajectories with high consistency. Quantitative results in Table~\ref{tab:autoreg_real_metrics} show that all joints achieve strong trajectory agreement, with angle $R^2$ values exceeding 0.92. 
	
	Notably, although the bucket joint exhibits relatively lower accuracy in angular velocity prediction ($R^2=0.5828$) due to its broader dynamic range, the integrated angle trajectory remains highly consistent ($R^2=0.9706$), indicating that high-frequency prediction errors do not accumulate significantly over time. These results confirm that the learned surrogate can reproduce the intrinsic dynamic behavior of the physical excavator under free-space motion.
	
	\subsubsection{Compound Operation Validation}
	
	To further evaluate generalization, 33 datasets (84k samples) are collected under compound operating conditions involving load interaction. The model is evaluated over 2950 autoregressive steps (295\,s at 10\,Hz); see Table~\ref{tab:exp_overview} for details.
	
	Initial experiments reveal that standard training leads to trajectory drift, particularly in the swing joint, due to systematic bias in angular velocity prediction. To address this issue, the bias-penalization term in the loss function (Section~\ref{sec:lstm}) is activated ($H=10$, $\lambda_{\text{bias}}=0.3$), effectively suppressing long-horizon drift.
	
\begin{figure*}[!htbp]
		\centering
		\subfloat[No-load condition]{%
			\includegraphics[width=0.48\textwidth]{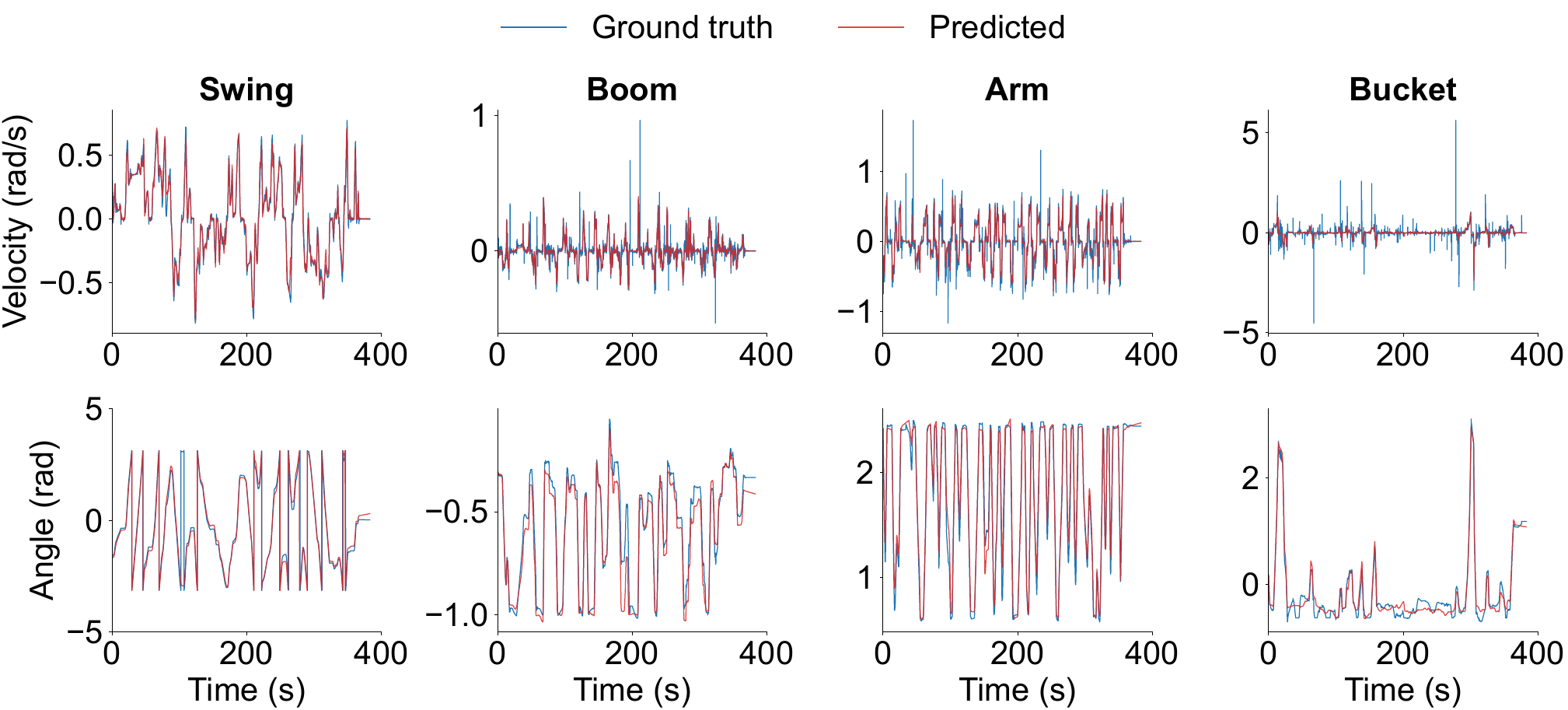}
			\label{fig:autoreg_real}
		}
		\hfill
		\subfloat[Compound condition]{%
			\includegraphics[width=0.48\textwidth]{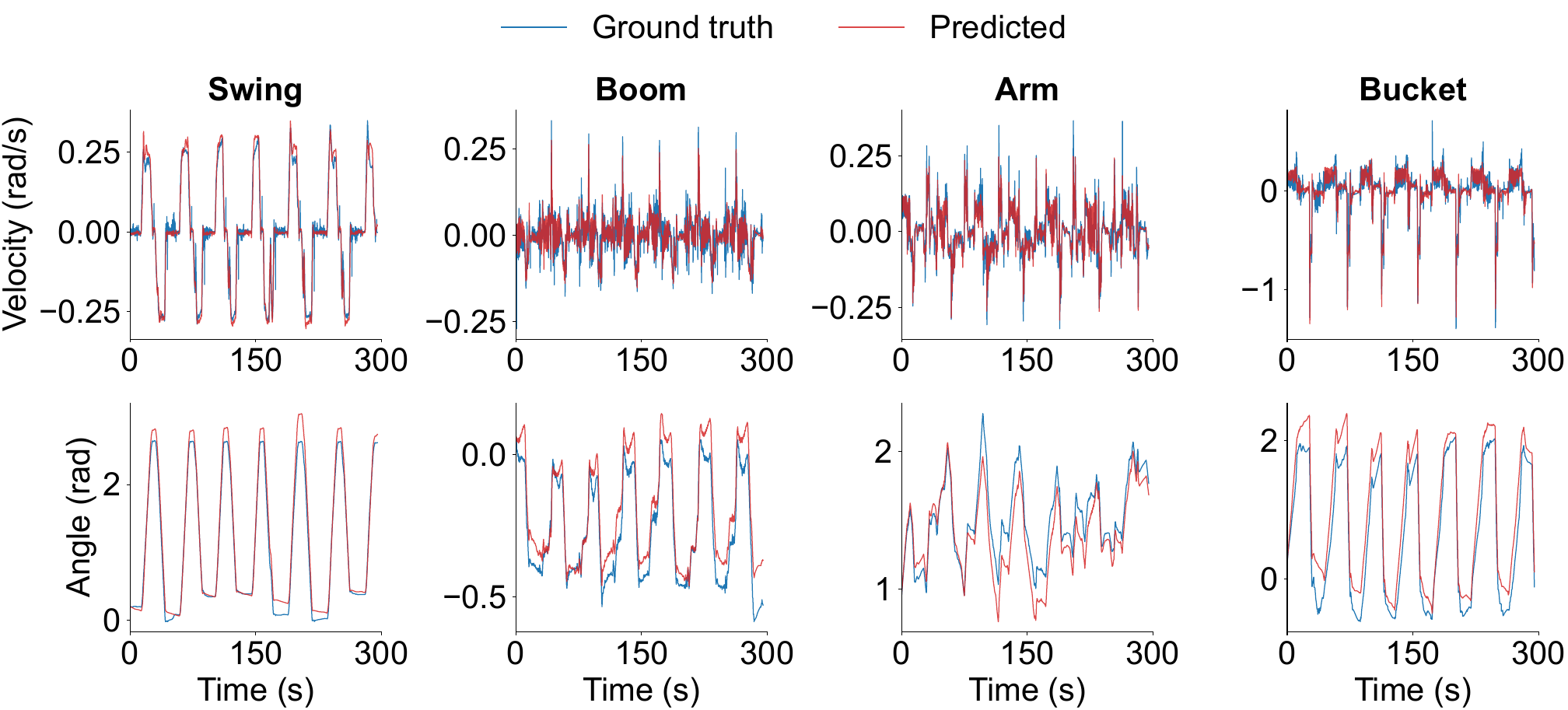}
			\label{fig:autoreg_real_compound}
		}
		\caption{Closed-loop validation on the physical excavator. Comparison between LSTM free-evolution trajectories and real system responses under identical control inputs.}
		\label{fig:autoreg_real_combined}
	\end{figure*}

	As shown in Fig.~\ref{fig:autoreg_real_compound}, the corrected model maintains stable trajectory alignment with the physical system under complex operations. Quantitative results in Table~\ref{tab:autoreg_real_metrics} indicate improved consistency across most joints compared to the no-load case, demonstrating that richer operational data enhances model generalization.

%
%

	\begin{table}[htbp]
		\centering
		\caption{Stage II: Closed-loop autoregressive validation metrics on the real excavator (No-load / Compound).}
		\label{tab:autoreg_real_metrics}
		\resizebox{0.94\columnwidth}{!}{%
			\begin{tabular}{c l c c c}
				\toprule
				\textit{Type} & \textit{Joint} & \textit{RMSE} & \textit{MAE} & \textit{$R^2$} \\
				\midrule
				\multirow{4}{*}{\textit{\makecell{Angular \\velocity\\ $\omega$}}}
				& Swing  & 0.0481 / 0.0249 & 0.0302 / 0.0175 & 0.9755 / 0.9750 \\
				& Boom   & 0.0342 / 0.0259 & 0.0164 / 0.0171 & 0.8487 / 0.7920 \\
				& Arm    & 0.0941 / 0.0269 & 0.0466 / 0.0185 & 0.8841 / 0.8734 \\
				& Bucket & 0.1528 / 0.0753 & 0.0559 / 0.0516 & 0.5828 / 0.8815 \\
				\midrule
				\multirow{4}{*}{\textit{\makecell{Angle\\ trajectory\\ $\theta$}}}
				& Swing  & 0.1822 / 0.1451 & 0.1553 / 0.1120 & 0.9989 / 0.9774 \\
				& Boom   & 0.0691 / 0.0763 & 0.0496 / 0.0654 & 0.9286 / 0.8210 \\
				& Arm    & 0.1053 / 0.1469 & 0.0733 / 0.1251 & 0.9737 / 0.7309 \\
				& Bucket & 0.1200 / 0.3395 & 0.0984 / 0.3101 & 0.9706 / 0.8667 \\
				\bottomrule
				\multicolumn{5}{l}{\footnotesize Each entry is reported as No-load / Compound.}
			\end{tabular}%
		}
	\end{table}
	
	\subsection{Discussion}
	
	Across both simulation and real-world experiments, the proposed surrogate model consistently reproduces the system-level behavior of the excavator under identical control inputs. The simulation results validate the correctness of the operator learning paradigm under ideal conditions, while the real-world results demonstrate its robustness in the presence of hydraulic nonlinearities and sensor imperfections.
	
	The introduction of consistency-aware state estimation ensures reliable training data, and the bias-aware training strategy further improves long-horizon stability under complex operations. Overall, these results confirm that the learned LSTM model can serve as a functional surrogate for both the simulator and the physical excavator in closed-loop settings.

	\section{Conclusions}\label{sec:conclusions}

This paper presents a simulation-to-real framework for learning system-level digital surrogates of hydraulic excavators. By formulating the excavator as an input–output operator, the proposed approach bypasses explicit dynamics modeling and directly learns the closed-loop behavior using an LSTM network. A consistency-aware state estimation method is introduced to address measurement inconsistencies in real-world data, enabling stable and physically meaningful training. Through staged validation in both simulation and physical systems, the learned surrogate demonstrates strong closed-loop behavioral equivalence, accurately reproducing both short-term responses and long-horizon trajectories.

These results establish that a data-driven surrogate can serve as a functional replacement for both simulators and physical machines in control and learning pipelines. Future work will focus on leveraging the surrogate for large-scale policy learning and improving robustness under broader operating conditions.

\end{document}